\documentclass{article}

\usepackage{arxiv}

\usepackage[utf8]{inputenc} 
\usepackage[T1]{fontenc}    
\usepackage{hyperref}       
\usepackage{url}            
\usepackage{booktabs}       
\usepackage{amsfonts,amsmath}       
\usepackage{nicefrac}       
\usepackage{microtype}      
\usepackage{lipsum}
\usepackage{graphicx}
\usepackage{bm}
\graphicspath{ {./images/} }

\title{A physics-informed machine learning model for reconstruction of dynamic loads}

\author{
 Gledson Rodrigo Tondo$^{\star}$ \\
  Bauhaus-Universität Weimar\\
  Weimar, Germany \\
   \And
 Igor Kavrakov \\
  University of Cambridge\\
  Cambridge, United Kingdom \\
  \And
 Guido Morgenthal \\
  Bauhaus-Universität Weimar\\
  Weimar, Germany \\
}

\begin{document}
\maketitle
\begin{abstract}
Long-span bridges are subjected to a multitude of dynamic excitations during their lifespan. To account for their effects on the structural system, several load models are used during design to simulate the conditions the structure is likely to experience. These models are based on different simplifying assumptions and are generally guided by parameters that are stochastically identified from measurement data, making their outputs inherently uncertain. This paper presents a probabilistic physics-informed machine-learning framework based on Gaussian process regression for reconstructing dynamic forces based on measured deflections, velocities, or accelerations. The model can work with incomplete and contaminated data and offers a natural regularization approach to account for noise in the measurement system. An application of the developed framework is given by an aerodynamic analysis of the Great Belt East Bridge. The aerodynamic response is calculated numerically based on the quasi-steady model, and the underlying forces are reconstructed using sparse and noisy measurements. Results indicate a good agreement between the applied and the predicted dynamic load and can be extended to calculate global responses and the resulting internal forces. Uses of the developed framework include validation of design models and assumptions, as well as prognosis of responses to assist in damage detection and structural health monitoring.
\end{abstract}


\section{Introduction}
Assumptions on statistical wind properties, limitations of aerodynamic models and restrictions in the stochastic dynamic analysis are just a few of many sources of uncertainties when modelling aerodynamic loads during the design phase of a long-span bridge. Furthermore, due to its extended lifetime and external effects such as climate change, the dynamic forces that are considered during design are subject to unforeseen changes. These factors motivate the creation of models to reconstruct dynamic loads based on measurement data.

Several methods for force reconstruction exist in the literature, which are generally based on data-driven techniques~\cite{liu2021dynamic,liu2022artificial}, optimization strategies~\cite{wang2015efficient} and defined basis functions~\cite{he2018adaptive}. A review of several of these models is given in~\cite{sanchez2014review}. A novel methodology based on stochastic processes is proposed in this study. The framework combines data-driven models with physics-based formulations, overcomes the necessity of regularization, naturally incorporates measurement noise properties, and can integrate different data types and sets with different measurement quality.

This study employs a specific aerodynamic model, based on the quasi-steady assumption, to evaluate the structural response of the Great Belt East Bridge due to an applied wind load. The evaluated response is contaminated with noise and further used as input to the physics-informed machine learning model, which in turn yields a stochastic model for the underlying aerodynamic force. Comparisons and evaluation of the results are provided by comparing the true and reconstructed signals, and the predictions are coupled with a structural finite element model to provide insights on structural responses and internal forces due to the aerodynamic loading.

\section{Gaussian process for force reconstruction}

\subsection{Physics-informed Gaussian process} \label{sec:sec21}

\begin{figure}[h]
  \centering
  \includegraphics[width=0.99\textwidth]{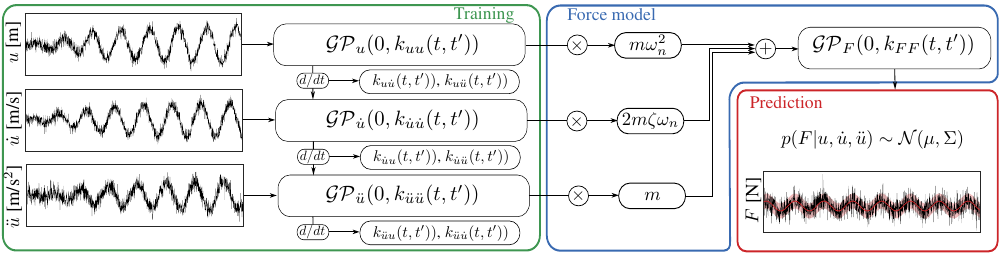}
  \caption{Framework for the physics-informed Gaussian process: (green) models for measured data are created and jointly trained. A force model (blue) is built combining the previous models and the harmonic oscillator’s differential equation, which can be used for predictions (red).}
  \label{fig:fig1}
\end{figure}

The response of a harmonic oscillator to an arbitrary dynamic loading $F$ is given by the second-order inhomogeneous differential equation

\begin{equation}
    m \Ddot{u} + 2 m \zeta  \omega_n \dot{u} + m \omega_n^2 u = F,
    \label{eq:eq1}
\end{equation}

where $m$ is the oscillator’s mass, $\zeta$ is the damping ratio to critical, $\omega_n$ is the oscillator’s circular natural frequency, and $u$, $\dot{u}$ and $\Ddot{u}$ are the displacement, velocity, and acceleration responses, respectively. These responses are generally directly measurable through sensor devices, and therefore the displacement response can be modelled as

\begin{equation}
    u = f(t) + \epsilon,
\end{equation}

where $t$ is the time and $\epsilon = \mathcal{N} (0,\sigma_{n,u}^2 )$ the Gaussian noise in the measurement system, characterized by a variance $\sigma_{n,u}^2$. The same principle of a stochastic process applies to the velocity and acceleration signals, where each of them has a particular noise variance $\sigma_{n,\dot{u}}^2$ and $\sigma_{n,\Ddot{u}}^2$.  Assuming the underlying deflection response is a stochastic zero-mean Gaussian process, a model can be created such that 

\begin{equation}
    u(t) \sim \mathcal{GP} \left( 0, k_{uu} (t, t'; \sigma_s, \ell) + \sigma_{n,u}^2 \delta(t, t') \right),
\end{equation}

where $\delta$ is the Kronecker-Delta operator and $k_{uu}$ is a covariance kernel parametrized by a standard deviation amplitude $\sigma_s$ and a length scale $\ell$~\cite{rasmussen2006gaussian}. Because the displacement response is assumed to be continuous and smooth in time, $k_{uu}$ is herein modelled by the squared exponential (SE) kernel

\begin{equation}
    k_{uu} = \sigma_s^2 \mathrm{exp} \left( - \frac{1}{2} \left( \frac{t - t'}{\ell} \right)^2 \right),  
\end{equation}

which reflects the assumption that similar time indexes should have similar displacement responses. Since velocities and accelerations are time-derivatives of the displacements and exploiting the fact and any linear operation to a Gaussian process result in another GP, physics-informed models can also be created for velocities $k_{\dot{u} \dot{u}}$ and accelerations $k_{\Ddot{u} \Ddot{u}}$, as well the respective cross-covariances between all measurement types. The generated models are used in combination with the measurement data for training, as shown in Figure~\ref{fig:fig1} (green block).

Combining the derived response kernels with the oscillator model from Eq.~\ref{eq:eq1} gives rise to a physics-informed cross-covariance model between the dynamic load and the deflection,

\begin{equation}
    k_{Fu} (t, t'; \sigma_s, \ell) = m \frac{d^2}{dt^2} k_{uu} + 2m \zeta \omega_n  \frac{d}{dt} k_{uu} + m \omega_n^2 k_{uu},  
\end{equation}

while the models for the remaining responses are generated similarly. The force covariance is calculated by applying Eq.~\ref{eq:eq1} to the second time index of $k_{Fu}$, yielding

\begin{equation}
k_{FF} (t, t'; \sigma_s, \ell) = m \frac{d^2}{dt^2} k_{Fu} + 2m \zeta \omega_n  \frac{d}{dt} k_{Fu} + m \omega_n^2 k_{Fu},
\end{equation}

which can be used for predictions, as seen in Figure~\ref{fig:fig1} (blue and red blocks, respectively).

\subsection{Optimisation of model parameters}

Although Gaussian processes are generally regarded as non-parametric models, the choice of covariance kernel and the noise assumptions lead to free model parameters that shall be identified based on training data. Assuming measurements from deflections, velocities and accelerations are available, the parameter set to be identified is defined by $\bm{\theta}=\lbrace \sigma_s,\ell,\sigma_{n,u},\sigma_{n,\dot{u}},\sigma_{n,\Ddot{u}} \rbrace$, where the measurement noise can be disregarded if no data from a specific quantity is collected, or extended if multiple sets of data with different properties are available~\cite{tondoPhysicsinformedGaussianProcess2022}. Parameter identification is carried out via maximum likelihood estimation,

\begin{equation}
\bm{\theta}_{\mathrm{opt}} = \mathrm{argmax}_{\bm{\theta}} \ \mathrm{log} \ p(\bm{y}|\bm{t},\bm{\theta}) = \mathrm{argmax}_{\bm{\theta}} - \frac{1}{2} \bm{y}^T \bm{K}^{-1} \bm{y}^T - \frac{1}{2} \mathrm{log} \  |\bm{K}| - \frac{N}{2} \mathrm{log} 2 \pi,
\label{eq:eq7}
\end{equation}

where $N$ is the number of data points, $\bm{y} = \left[ \bm{u}, \dot{\bm{u}}, \Ddot{\bm{u}} \right]^T$  collects the training data and $\bm{K}$ is a block-covariance matrix calculated via the kernel formulations derived in Sec.~\ref{sec:sec21},

\begin{equation}
    \bm{K} = 
    \begin{bmatrix}
        k_{uu} (\bm{t}, \bm{t}') + \sigma_{n,u}^2 \delta (\bm{t}, \bm{t}') & k_{u\dot{u}} (\bm{t}, \bm{t}') & k_{u\Ddot{u}} (\bm{t}, \bm{t}') \\
        k_{\dot{u}u} (\bm{t}, \bm{t}') & k_{\dot{u}\dot{u}} (\bm{t}, \bm{t}') + \sigma_{n,\dot{u}}^2 \delta (\bm{t}, \bm{t}') & k_{\dot{u}\Ddot{u}} (\bm{t}, \bm{t}') \\
        k_{\Ddot{u}u} (\bm{t}, \bm{t}') & k_{\Ddot{u}\dot{u}} (\bm{t}, \bm{t}') & k_{\Ddot{u}\Ddot{u}} (\bm{t}, \bm{t}') + \sigma_{n,\Ddot{u}}^2 \delta (\bm{t}, \bm{t}')
    \end{bmatrix}.
    \label{eq:eq8}
\end{equation}	

for the specific measurement times $\bm{t}$. Maximization of Eq.~\ref{eq:eq7} is achieved by gradient ascent via quasi-Newton BFGS optimization. Although not explicitly shown, it is worth noting that the derived formulation does not require a regular time-step interval, providing flexibility in cases of missing response time steps or imperfect sensor synchronization.

\subsection{Force prediction}

Once the free parameters are identified based on data, the force signal that generated the measured responses can be reconstructed. The joint distribution of the data collected in $\bm{y}$ and the force vector $\bm{F}$ is given by

\begin{equation}
p(\bm{y},\bm{F}) = \mathcal{N} \left( 
\begin{bmatrix}
0 \\
0
\end{bmatrix}, \begin{bmatrix}
\bm{K} & \bm{K}_{\star} \\
\bm{K}_{\star}^T & \bm{K}_{\star \star}
\end{bmatrix} \right),
\end{equation}

with $\bm{K}_{\star} = \left[ k_{Fu} (\bm{t}_{\star}, \bm{t}'), k_{F \dot{u}} (\bm{t}_{\star}, \bm{t}'), k_{F \Ddot{u}} (\bm{t}_{\star}, \bm{t}')\right]^T$ and $\bm{K}_{\star \star} = k_{FF} (\bm{t}_{\star}, \bm{t}_{\star}')$  where $\bm{t}_{\star}$ are the time instants for force prediction. Conditioning the force model on the measurement data yields 

\begin{equation}
p(\bm{F} | \bm{y}) = \mathcal{N} (\bm{\mu} = \bm{K}_{\star}^T \bm{K}^{-1} \bm{y}, \bm{\Sigma} = \bm{K}_{\star \star} - \bm{K}_{\star}^T \bm{K}^{-1} \bm{K}_{\star})
\label{eq:eq10}
\end{equation}

which defines the prediction’s mean values $\bm{\mu}$ and covariance matrix $\bm{\Sigma}$ shown in Figure~\ref{fig:fig1}.

\section{Fundamental studies}

\begin{figure}[h]
  \centering
  \includegraphics[width=0.49\textwidth]{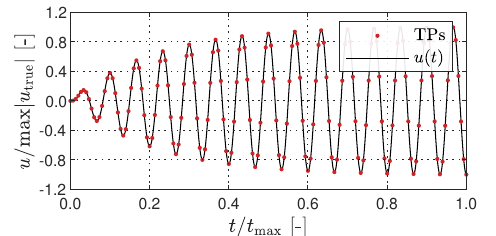}
  \includegraphics[width=0.49\textwidth]{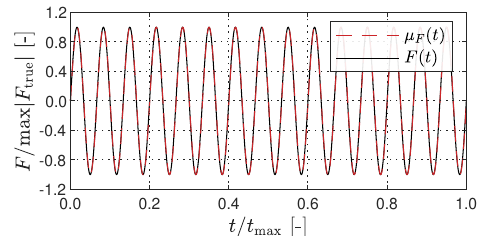}
  \caption{Left: displacement signal and corresponding training points (TPs). Right: true harmonic force and the model prediction’s mean. The prediction standard deviation is not shown and tends to zero as the training data contains no noise. }
  \label{fig:fig2}
\end{figure}

To evaluate the model’s performance in a controlled and simplified manner, a single-degree-of-freedom (SDOF) system response subjected to a harmonic force of unit amplitude is numerically calculated. The oscillator is set with a mass $m=1$~kg, a damping ratio of $\zeta=0.05$ and a circular natural frequency $\omega_n = 2 \pi$~rad/s. The sampling rate of the signal is defined by $f_s=200$~Hz, and the forcing signal is in resonance with the oscillator. Training data in $\bm{y}$ consist of sparse, regularly distributed and noise-free displacement readings, while velocity and acceleration responses are not provided to the model, leading effectively to only two optimizable parameters $\bm{\theta} = \lbrace \sigma_s, \ell \rbrace $. The complete displacement response and the training points (TPs) provided to the model are shown in Figure~\ref{fig:fig2} (left).	

After the optimal parameters are identified, predictions of the original forcing signal are made using Eq.~\ref{eq:eq10}. The true harmonic force and the probabilistic prediction results are shown in Figure~\ref{fig:fig2} (right). Under idealized conditions, a very good agreement is observed between the true and the predicted signal, in a mean ($\bm{\mu}_F$) sense. Due to the lack of noise in the training data, represented in the GP model by setting $\sigma_{n,u}=0$~m, the standard deviation $\sigma_F$ of the probabilistic force model tends to zero, indicating full model confidence in the predictions.
  
\begin{figure}[h]
  \centering
  \includegraphics[width=0.49\textwidth]{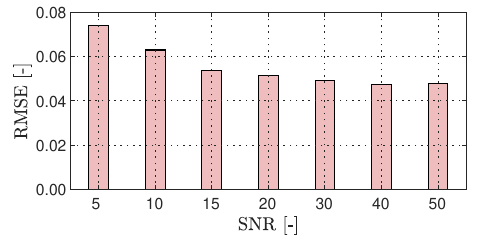}
  \includegraphics[width=0.49\textwidth]{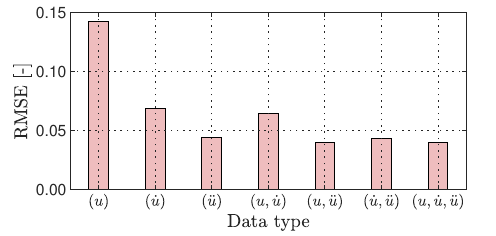}
  \caption{Root mean squared errors of (right) models with different SNRs, for a fixed data set $\bm{y} = \left[ \bm{u}, \dot{\bm{u}}, \Ddot{\bm{u}} \right]^T$, and (left) models trained with different data types, for a fixed SNR~=~20.}
  \label{fig:fig3}
\end{figure}

In contrast to the previous assumptions, measurement data is generally contaminated by noise originating from several possible different sources. The signal-to-noise ratio (SNR) is commonly employed to quantify the amount of noise in a given signal, herein defined as SNR~=~$A_\mathrm{signal}⁄A_{\mathrm{noise}}$, where $A$ is the root mean square function. Moreover, measurement data from all response types is generally not available, as it is in many cases redundant and not cost-effective. By construction, the GP model can nevertheless work with missing measured data or incorporate different multi-fidelity sets of the same data type~\cite{tondoPhysicsinformedGaussianProcess2022}. To evaluate the effects of both properties, the system is again trained for data sets containing various SNRs and composed of different combinations of data types. 

Although the GP formulation accounts for noise in the measured data via the $\sigma_n$ parameters, high-noise signals (low SNRs) can locally modify the force predictions, effectively shifting the mean prediction in a smoothed region around specific training points, leading ultimately to a high root mean squared error RMSE. The prediction performance increases rapidly, however, for increasing SNRs, indicating the influence of sensor quality on the regressed forces, as seen in Figure~\ref{fig:fig3} (left). Differences between predicted and true force signals also reduce when more than one data type is available for training, as shown in Figure~\ref{fig:fig3} (right). For cases when measurements contain noise, providing different data types allows the model to find a balancing point that explains the full measurement, represented by the covariance kernel in Eq.~\ref{eq:eq8}, leading to a stabilized prediction. In addition, Figure~\ref{fig:fig3} (right) indicates that providing acceleration data for training leads to better performance when compared to velocity and displacement measurements. While this is the case for the particular SDOF oscillator considered in this comparison, results are also a function of the system and applied force properties, and therefore cannot be generalized.

\section{Aerodynamic forces on the Great Belt East Bridge}

\begin{figure}[h]
  \centering
  \includegraphics[width=0.99\textwidth]{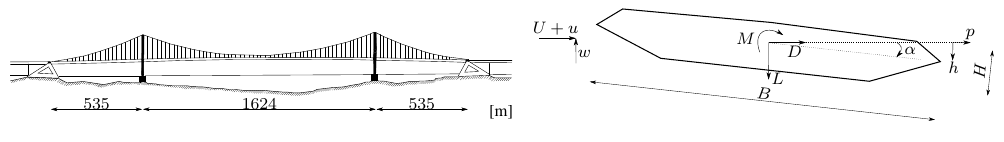} \\
  \includegraphics[width=0.49\textwidth]{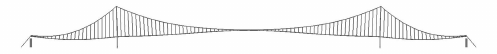}
  \includegraphics[width=0.49\textwidth]{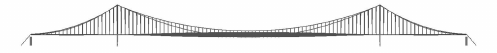}
  \caption{The Great Belt East Bridge. On top, the elevation (left) and cross-section sketch with the coordinate system for aerodynamic forces and wind fluctuations (right). At the bottom, the first vertical ($f_h=0.100$~Hz, left) and torsional ($f_\alpha=0.278$~Hz, right) mode shapes.}
  \label{fig:fig4}
\end{figure}

The developed model for force identification is employed to reconstruct the aerodynamic forces applied to a numerical model of the Great Belt East Bridge in Denmark. The suspension bridge has a main span of 1624 m and two symmetrical side spans of 535 m each. The deck has a streamlined cross-section with a chord $B=31$~m and depth $H=4$~ m. A sketch of the bridge elevation and the coordinate system for the wind fluctuations and aerodynamic forces on the cross-section level, for the reduced-order dynamic modelling, are shown in Figure~\ref{fig:fig4}, along with the first vertical and torsional mode shapes. 
 
For the aerodynamic analysis of the bridge, the wind velocity is separated into a mean component $U$ and fluctuating components $u$ and $w$ in the horizontal and vertical directions, respectively. The 2D section is assumed to have three degrees of freedom for horizontal $p$, vertical $h$ and rotational $\alpha$ motions. The drag $D$, lift $L$ and moment $M$ forces acting on the deck due to motion, mean wind and buffeting are obtained using the quasi-steady assumption~\cite{chen2001nonlinear} by

\begin{align}
    D = F_L \ \mathrm{sin} \ \phi_D - F_D \ \mathrm{cos} \ \phi_D, & & L = F_L \ \mathrm{cos} \ \phi_L - F_D \ \mathrm{sin} \ \phi_L, & & M = F_M,
\end{align}

with $\phi_i$ for $i=\lbrace D,L,M \rbrace$ being the dynamic angle of attack, and

\begin{align}
    F_D = \frac{1}{2} \rho U_{rD}^2 B C_D (\alpha_{eD}), & & F_L = \frac{1}{2} \rho U_{rL}^2 B C_L (\alpha_{eL}), & & F_M = \frac{1}{2} \rho U_{rM}^2 B^2 C_M (\alpha_{eM})
\end{align}

where $\rho$ is the air density, $C_i$ is the static wind coefficient and $\alpha_e$ the effective angle of attack, calculated by 

\begin{equation}
    \alpha_{ei} = \alpha_s + \alpha + \phi_i = \alpha_s + \alpha + \mathrm{arctan} \left( \frac{w + \dot{h} + m_i B \dot{\alpha}}{U + u - \dot{p}} \right)
\end{equation}

where $\alpha_s$ is the angle of attack at static equilibrium. The resultant velocity $U_{ri}$ is calculated as

\begin{equation}
    U_{ri} = \sqrt{ \left( U + u - \dot{p} \right)^2 + \left( w + \dot{h} + m_i B \dot{\alpha}\right)^2 }
\end{equation}

where $m_i$ is the aerodynamic centre for $i=\lbrace D,L,M \rbrace$. The dynamic structural response was calculated for a 10-minute turbulent wind time history. The mean wind speed considered is $30$~m/s, while the isotropic turbulence is defined by an intensity of 10\% and generated using the von Kármán spectrum, with length scales of 200~m for the horizontal direction and 100~m for vertical and longitudinal directions, respectively. Mechanical admittance in the buffeting response was calculated using Sears’ model and the aerodynamic centre was defined using Den Hartog’s assumptions. The static wind coefficients are obtained from~\cite{kavrakov2018synergistic}.
   
\begin{figure}[h]
  \centering
  \includegraphics[width=0.3275\textwidth]{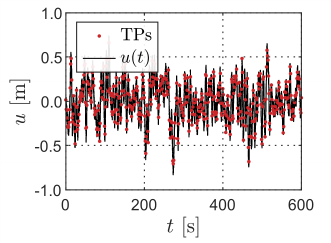}
  \includegraphics[width=0.3275\textwidth]{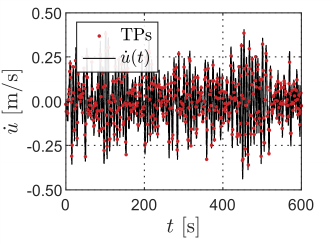}
  \includegraphics[width=0.3275\textwidth]{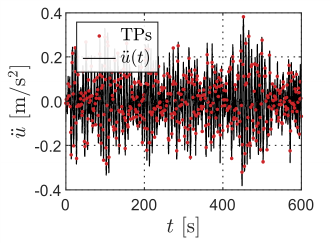}\\
  \includegraphics[width=1\textwidth]{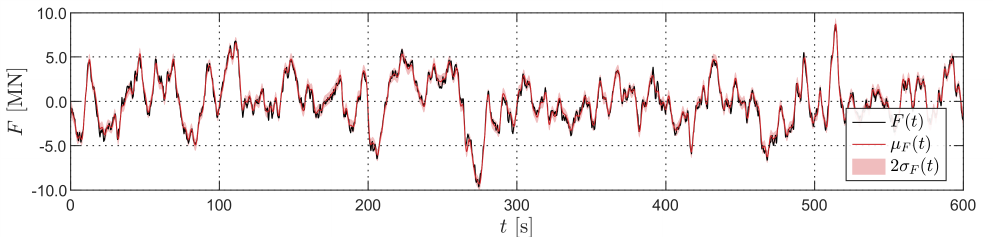}
  \caption{Top: noise-contaminated displacement, velocity, and acceleration responses from the first vertical mode and corresponding training points (TPs). Bottom: true dynamic force and the prediction’s mean and 95\% confidence interval.}
  \label{fig:fig5}
\end{figure}

\begin{figure}[h]
  \centering
  \includegraphics[width=0.495\textwidth]{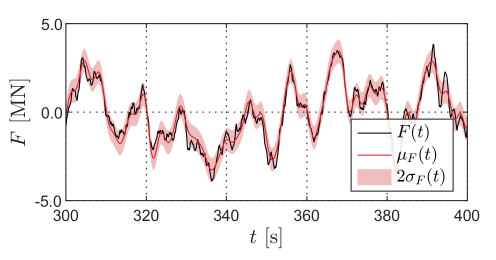}
  \includegraphics[width=0.495\textwidth]{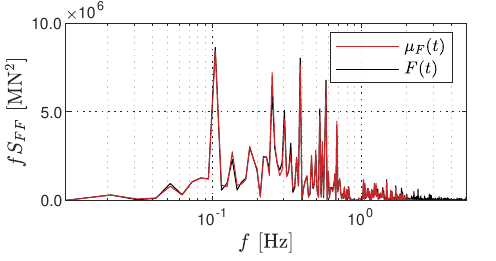}
  \caption{Left: detail of the probabilistic prediction in a smaller time window. Right: PSD of the original and regressed mean modal force.}
  \label{fig:fig6}
\end{figure}

A total of 22 modes are considered in the aerodynamic analysis, and the modal responses are linearly combined to yield the global coordinate response. Herein, a white noise signal of SNR = 20 is added to each of the modal responses, and the resulting time series is directly used to obtain the forcing signal. In practice, sensor measurements in global coordinates can be decomposed into modal components if modal information is available, using techniques such as modal decomposition or Kalman filtering~\cite{nonomura2018dynamic}. A sparse selection of the displacement, velocity and acceleration responses is used for training, where the data points were selected using a regular time interval of $\Delta t= 1.25$~s (cf. Figure~\ref{fig:fig5}, top). The resulting prediction of the modal force for the whole 10-minute analysis time, in the case of the first vertical mode, is shown in Figure~\ref{fig:fig5}, bottom.
Similar to the SDOF results, a good agreement is observed between the true modal force and the GP predictions, even though in the bridge model the force signal contains multiple harmonic components, governed typically by the structural modal frequencies. If a shorter time window is analysed (cf. Figure~\ref{fig:fig6}, left), however, it is evident that the predictions are a smoothened version of the true force, with the GP effectively acting as a low-pass filter of the original signal and capturing high-frequency content by its uncertainty measurement. This is also verified by the power spectral density $S_{FF}$ of the force, as shown in Figure~\ref{fig:fig6} (right), where a good agreement is observed until $f=2.5$~Hz, corresponding approximately to the Nyquist frequency related to the sampling rate of the data used for training. 

Evaluation of the prediction quality for all 22 modes is carried out using the mean squared error in relation to the normalized true force. Although the modal contributions towards the global response are functions of the forcing and structural properties, similar RMSE values between 0.04 and 0.10 are observed across all modes, and the predictions have similar quality levels. In practice, this may not always be true due to the influence of noise in the measurement system.
 
\begin{figure}[h]
    \centering
    \includegraphics[width=1\textwidth]{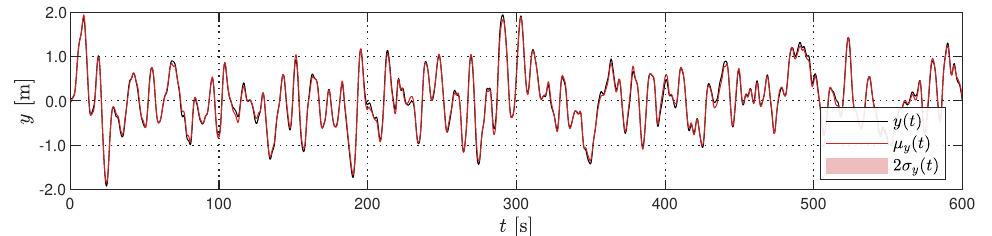}\\
    \includegraphics[width=1\textwidth]{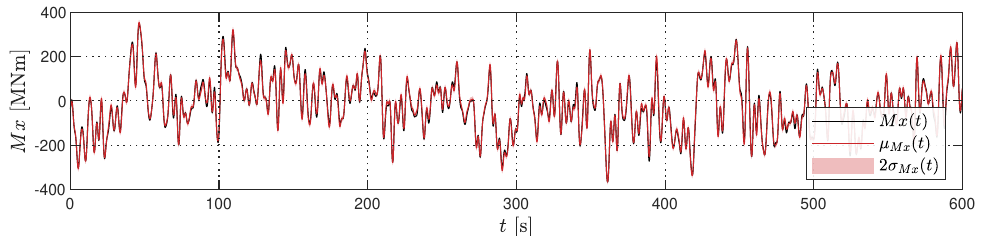}
    \caption{Comparison between numerical response and probabilistic predictions. Top: deck global vertical response at midspan. Bottom: internal bending moment response at pylon support. }
    \label{fig:fig7}
\end{figure}

The force predictions can further be used to evaluate global responses based on the measurement data, by combining the GP predictions with a finite element model of the Great Belt East Bridge. Hence, two different examples are now presented. In Figure~\ref{fig:fig7} (top), the global vertical displacement response at midspan is reconstructed based on the superposition of modal displacements, in combination with the corresponding mode shapes. In Figure~\ref{fig:fig7} (bottom), the bending moment values at the pylon support are shown based on the numerical analysis and on the GP predictions. The model uncertainty is reduced in comparison to the force predictions since the structural response is generally governed by lower frequencies, which are captured with higher accuracy by the Gaussian process model, as observed in Figure~\ref{fig:fig6} (right). Furthermore, a good agreement is observed between the true response and the mean predictions.

\section{Summary and conclusions}

In summary, a novel stochastic method to reconstruct dynamic forces based on sensor measurement data has been presented. The model is built based on Gaussian process regression for machine learning, and relations between physical properties are embedded using well-established differential equations. This hybrid formulation allows for the use of heterogeneous and multi-fidelity data during training. Regularization schemes, generally a source of problems in typical optimization problems, are bypassed by the natural trade-off between data fitting and model complexity provided by the Gaussian likelihood, which also allows for fully analytical tractability for posterior sampling.

Force predictions were compared and discussed from the perspective of the assumptions taken for each particular example. The model limitation to represent a wide range of frequency content correlates with the sampling rate of training data, which may prove problematic when high-frequency components are expected, as the optimization for GPs scales particularly poorly for an increasing number of training points. Cases of missing steps or unsynchronized training data can be seamlessly incorporated by the derived model but were not considered in this study. Moreover, coupling the probabilistic force predictions with a finite element solver allows for a statistical view of global responses and, consequently, internal forces. The outputs of this analysis provide valuable insights into structural performance evaluation and allow for condition diagnostics and prognosis.

In conclusion, the derived framework is a powerful statistical tool for dynamic force reconstruction and has direct applications in model validation and structural health monitoring. Although the results presented herein are for a specific case study, it is expected that they extend to different structural systems and dynamic loading properties.

\bibliographystyle{unsrt}  
\bibliography{references}

\end{document}